\documentclass{article}
\usepackage{arxiv}
\usepackage{amsmath}
\usepackage[utf8]{inputenc} % allow utf-8 input
\usepackage[T1]{fontenc}    % use 8-bit T1 fonts
\usepackage{hyperref}       % hyperlinks
\usepackage{url}            % simple URL typesetting
\usepackage{booktabs}       % professional-quality tables
\usepackage{amsfonts}       % blackboard math symbols
\usepackage{nicefrac}       % compact symbols for 1/2, etc.
\usepackage{microtype}      % microtypography
\usepackage{lipsum}
\usepackage{float}
\usepackage{graphicx}
\usepackage{multirow}
\usepackage{array}
\usepackage{comment}
\usepackage{longtable}
\usepackage{ragged2e}
\usepackage{subfig}

\title{Driver Drowsiness Detection Using Ensemble Convolutional Neural Networks on YawDD}

\author{
Rais Mohammad Salman\\
Mechatronics Engineering\\
International Islamic University Malaysia\\
Kula Lumpur, 43200, Malaysia\\
\texttt{salmanmech2018@gmail.com}\\
\And
Mahbubur Rashid\\
Mechatronics Engineering\\
International Islamic University Malaysia\\
Kula Lumpur, 43200, Malaysia\\
\texttt{mahbub96@gmail.com}
\And
Rupal Roy\\
Mechatronics Engineering\\
International Islamic University Malaysia\\
Kula Lumpur, 43200, Malaysia\\
\texttt{rupal.roy@live.iium.edu.my}
\And
  Md Manjurul Ahsan \\
  Industrial and Systems Engineering\\
  University of Oklahoma\\
  Norman, Oklahoma-73071 \\
  \texttt{ahsan@ou.edu} \\
   \And
 Zahed Siddique \\
  School of Aerospace and Mechanical Engineering\\
  University of Memphis\\
  Tennessee, USA\\
  \texttt{zsiddique@ou.edu} \\
}

\begin{document}
\maketitle

\begin{abstract}
Driver drowsiness detection using videos/images is one of the most essential areas in today's time for driver safety. The development of deep learning techniques, notably Convolutional Neural Networks (CNN), applied in computer vision applications such as drowsiness detection, has shown promising results due to the tremendous increase in technology in the recent few decades. Eyes that are closed or blinking excessively, yawning, nodding, and occlusion are all key aspects of drowsiness. In this work, we have applied four different Convolutional Neural Network (CNN) techniques on the YawDD dataset to detect and examine the extent of drowsiness depending on the yawning frequency with specific pose and occlusion variation. Preliminary computational results show that our proposed Ensemble Convolutional Neural Network (ECNN) outperformed the traditional CNN-based approach by achieving an F1 score of 0.935, where other three CNN, such as CNN1, CNN2, and CNN3 approaches gained 0.92, 0.90, and 0.912 F1 scores, respectively.
\end{abstract}

% keywords can be removed

\section{Introduction}
\begin{figure*}
    \centering
    \includegraphics[width=\textwidth]{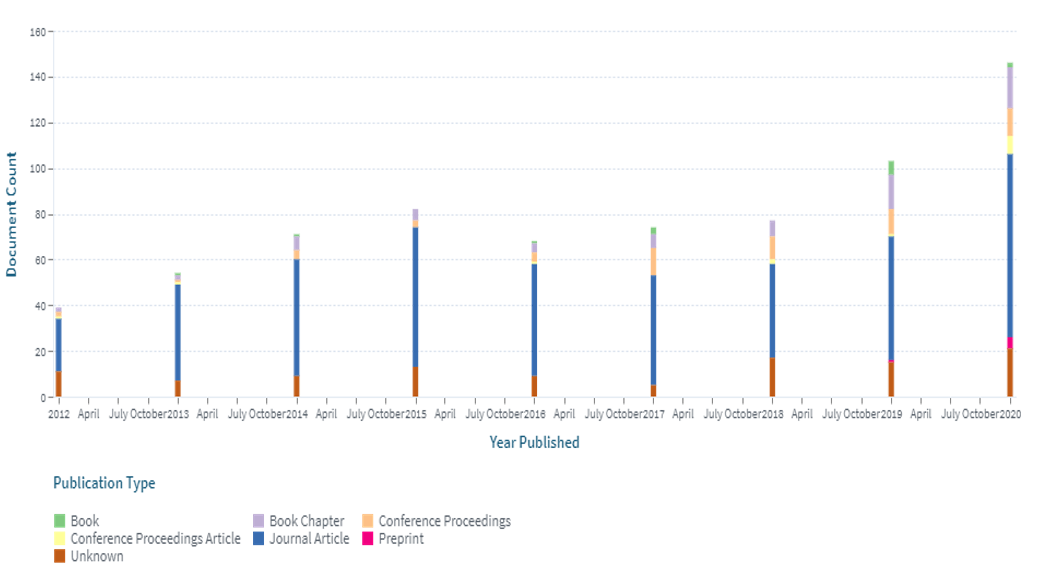}
    \caption{Scholarly works on drowsiness detection using deep learning from 2012 to 2020.}
    \label{drwasipubl}
\end{figure*}

In the last few decades, computer vision applications such as drowsiness detection for the safety and well-being of employers have been an important and concerning factor among researchers and is one of the hot topics in the field of Artificial Intelligence. With the advent of revolutionary, sophisticated, and human interactive machines and technologies, the research has been capable of developing real-time applications for human safety~\cite{ahsan2020face}. Detecting drowsiness by detecting behavioral features such as eyes, mouth, facial features~\cite{gupta2017robust}, etc., is one of the very important aspects considered by the researchers~\cite{ahsan2020face}. However, different approaches can be regarded as for driver drowsiness detection, such as vehicular-based approach (such as steering movement detection, force on acceleration pedal detection, etc.) and physiological-based approach (EEG, ECG, etc.)~\cite{ramzan2019survey}. A lot of work has been continuously carried out to improve drowsiness detection by enhancing the accuracy and precision of the model~\cite{abtahi2014yawdd,weng2016driver,revelo2019human}. The target or aim to driver drowsiness detection can be achieved effectively by using some of the techniques mentioned above such as neural network,CNN, and deep CNN which has shown remarkable results. Video-based driver drowsiness detection with deep learning techniques includes different parameters such as proper feature extractions, adequate computing power, and synchronized human-computer interactions. Additionally, behavioral feature extraction plays a significant role in drowsiness detection.\\
To develop a suitable video-based driver drowsiness detection model, proper facial frames must be extracted within the working domain~\cite{pan2007eyeblink, ramzan2019survey}. Using deep learning~\cite{ahsan2021detection, ahsan2021detecting, ahsan2020deep, ahsan2020covid} techniques instead of conventional machine learning approaches provides better results in terms of accuracy, precision, and prediction~\cite{manjurul2021machine}. Convolutional Neural Networks (CNN) is one of the good image classifier approaches because of its capability to detect spatial and temporal features in the image with the application of appropriate filters or kernels~\cite{zhou2012image}.
 Figure~\ref{drwasipubl} displays referenced literature in drowsiness detection using deep learning from 2012 to 2020. Different training architectures have been implemented by many researchers to improve the performance of drowsiness detection. 
The behavioral approach such as eye-tracking, yawning and face detection plays an important role in drowsiness detection~\cite{ramzan2019survey}. In the eye-tracking-based approach, Viola-Jones algorithm demonstrates faster performance compared to deep neural networks. However, when the object is moving, the performance of viola-jones algorithms in eye or face localizing significantly degraded~\cite{zhang2015multimodal}. A recent study conducted by Revelo el al. (2019) demonstrated that DNN and LSTM based approach performs well for yawning detection, where Spatio-temporal feature analysis was considered as one of the main factor~\cite{revelo2019human}. Xie et al. (2019) also used DNN approaches to detect the human drowsiness in real time. Their model showed robustness even in different situations, such as in the case of lack of illumination, positions, and view angles~\cite{xie2018real}. Some of the recent CNN based approaches such as AlexNet~\cite{krizhevsky2017imagenet}, VGG-FaceNet~\cite{parkhi2015deep}, Flow ImageNet~\cite{donahue2015long}, ResNet~\cite{he2016deep} also shows promising result in detecting drowsiness. However, the performance of those CNN artchitecture varies depending on the designated computer vision problems. For instance, AlexNet performs well compared to other models in different contexts and environmental shifts, such as indoor and outdoor, day and night;  VGG-FaceNet is performed well for the extraction of facial features such as racial groups. On the other hand, for behavioral characteristics and head movements, Flow ImageNet, and  for hand gestures, ResNet demonstarted better performance~\cite{park2016driver}.\\
Ensemble-based techniques, on the other hand, have recently gained much popularity due to their versatility in application. Typically, Ensemble approaches are used to train a "large" number of CNN models. Voting or averaging the total performance during the prediction is used to evaluate the performance. To begin, multiple neural networks are trained, with each network returning the likelihood of each class mark. Before the training phase, the models chosen to be Ensemble, which is loaded with their weights and architecture~\cite{ba2013deep}, those probabilities are weighted together, and the final classification is obtained~\cite{mohanraj2019ensemble}.
There has been substantial progress in behavioral feature recognition for driver drowsiness detection, just as there has been lot of progress in signal processing for physiological feature-based driver drowsiness detection. A lot of multimodel methods have been considered to solve these issues~\cite{zhang2015multimodal,choi2018driver}. In the last decades, there has been a lot of deep learning based research has been conducted in the field of computer vision on driver drowsiness detection to ensure the safety of the driver~\cite{ukwuoma2019deep}. 

\begin{figure*}[h!]
    \centering
    \includegraphics[width=\textwidth]{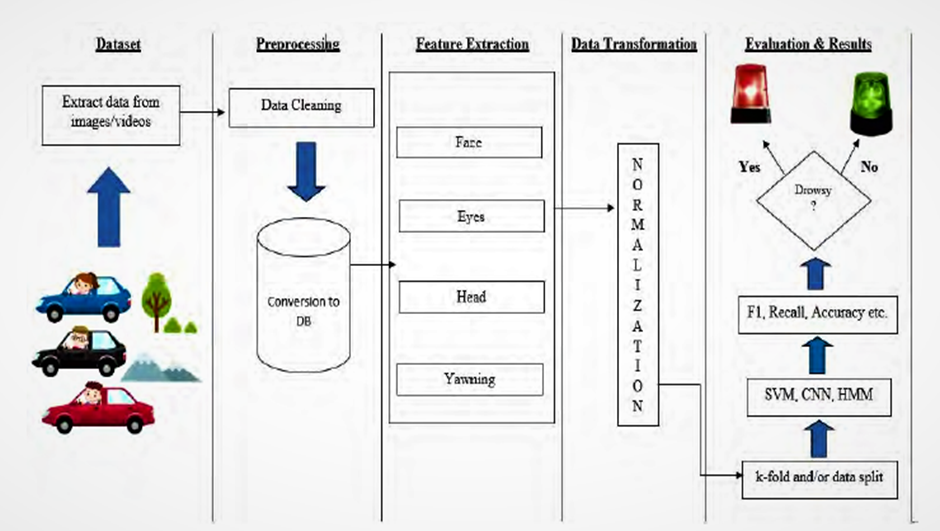}
    \caption{Pre and post processing approaches in video-based driver drowsiness detection for behavioral approaches.}
    \label{fig1}
\end{figure*}

This paper aims to develop a video-based driver drowsiness detection model using the deep learning technique on the YawDD dataset. However, video-based driver drowsiness detection is not a simple task, as demonstrated in Figure~\ref{fig1}, and requires multiple pre and post-processing approaches.\\

\section{Methodology}
 The driver's behavior for drowsiness detection is usually categorized as alert or drowsy or attentive or non-attentive. So, depending on these classes' images/frames are extracted from input video samples~\cite{nikolskaia2019prototype}. The number of frames to be extracted varies based on the designated problem statement, complexity of the work, computing ability and processing time. The frames extracted from the video database are then used for Face detection, which is the next step after pre-processing. It is one of the very common approaches that can detect drowsiness by facial features like yawning, eye blinking, etc. Hence, it is very important to detect faces to obtain facial features to move further for processing and recognition. Many face detection algorithms are being used by researchers, such as Viola-Jones, Local Binary Pattern Histograms (LBPH), Eigenfaces, Fisher face, OpenCV, and, DLIB~\cite{delbiaggio2017comparison,srivastava2017survey}. \\
 Conventional algorithms acknowledge the face by identifying and extracting facial highlights from the face image. For example, many researchers have been using facial landmarks (like 68 landmarks)~\cite{yuen2017occluded} to extract the facial key features, which calculates and examines the state of the eyes, mouth, etc. The extracted features would then be used to look for similar matching features of other images. In recent years the industries have moved and focused on deep learning techniques for different applications. CNN is a very strong performer, has been used for the calculation of acknowledging face detection. These calculations, which are achieved in the form of pixels from images, describe the arrangement of the picture information. These assimilated features include the PERCLOS (percentage of Eye Closure)~\cite{mandal2016towards,savacs2018real}, the ratio of mouth opening~\cite{kumar2018driver}, movement of the head or pose variation~\cite{lyu2018long}. CNN separates or distinguishes a huge number of these highlighted features from the image very well.\\
The cropping procedure comprises removing the portion of unwanted regions from the picture to remove incidental unnecessary features from the image during the face detection to improve its surroundings or highlight the regions (face) from its background. Once the cropping is done, the image resizes, and the size of the image varies and depends on the requirement. During this experiment we have resized all the image into 90x90 pixels to reduce the processing time while preserving the features or attributes of the image with this specific dimensions~\cite{zhou2012image}. Note that, the classification accuracy depends on how well the features are extracted during image pre-processing. Therefore, classification accuracy will automatically improve by optimizing the selected features~\cite{boubenna2018image}.

\subsection{CNN structure}
Convolutional neural network (CNN) is one of the most well-known deep learning (DL) algorithms on which, when an image is passed; it applies learnable weights and bias to a different point in the image; this way, it is able to separate features from one another. CNN requires less pre-processing compared to other DL algorithms. The design of CNN is inspired by the visual cortex of the human brain. Convolutional Net is an arrangement of different types of layers, and these layers of CNN perform different operations when input passes through these layers. The general architecture of CNN is described as follows:\\
\begin{itemize}
    \item CNN layers- Convolutional layers are the layers under which a deep CNN filter is added to the original image or other feature maps. A filter in a CNN is like a weight matrix which used to produce a convoluted output; we multiply a part of the input image. For example, let's assume we have got a size of 28*28 images. To form what is known as a convoluted output, we randomly allocate a filter of size 3*3, which is then multiplied with various 3*3 parts of the image.
\item ReLu- ReLu is an activation function also known as rectified linear activation unit. The main advantage of using ReLu is that it has a constant derivative value for all inputs greater than 0. The constant derivative value helps to train the network faster. Instead of sigmoids, the recent networks prefer using ReLu activation functions for the hidden layers. The function is defined as:
$f(x) = max(x,0)$

The output of the function is 1 when X>0 and 0 for X<=0. Figure~\ref{fig3} demonstrates the graphical representation of ReLu.  
\begin{figure}[h!]
    \centering
    \includegraphics[width=.5\textwidth]{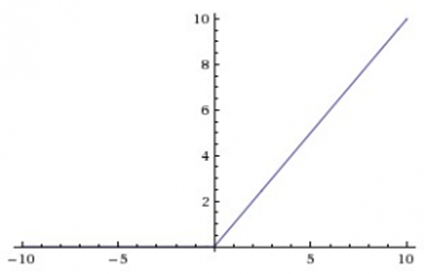}
    \caption{ReLu functions.}
    \label{fig3}
\end{figure}
\item Pooling layer –  The pooling layer is used to reduce the dimensionality of the network. The pooling layers are usually two types max-pooling (takes the maximum value in a particular filter region) and average pooling (takes the average value in a filter region). For example, the most common type of pooling is a pooling layer of filter size (2,2) using the MAX operation. What it would do is, it would take the maximum of each 4*4 matrix of the original image. Figure~\ref{fig4} displays the max pooling output with 2x2 filters when applied on 4*4 matrix.
\begin{figure}[h]
    \centering
    \includegraphics[width=.7\linewidth]{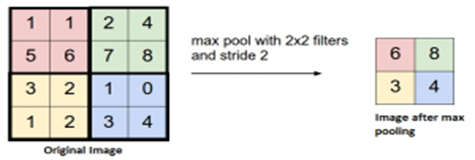}
    \caption{Example of max pooling with 2x2 filter.}
    \label{fig4}
\end{figure}
\item FC- FC also known as Fully connected layers, that are placed before the classification output of a CNN and are used to flatten the results before classification. Here in Figure~\ref{fig5} illustrates the flow diagram of CNN based approached applied on image classification.\\
\begin{figure*}
    \centering
    \includegraphics{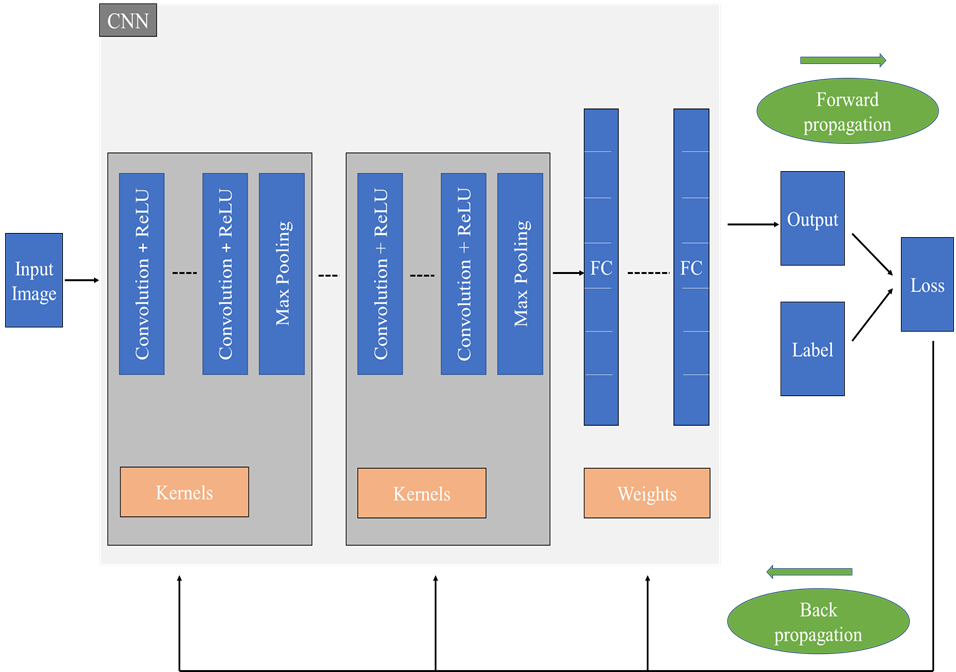}
    \caption{An schematic view of CNN based approach applied in this study.}
    \label{fig5}
\end{figure*}
\end{itemize}
\subsubsection{Convolutional Neural Net 1} Three convolutional layers, one pooling layer, two fully connected layers, and a softmax unit with polling applied to each layer to comprise the Convolutional Neural Net 1 (CNN1) architecture. The first, second and third layer contains 3x16, 3x32, filter, 3x64 filters respectively. Each layer is followed by rectified linear units (ReLU) and a 3x3 stride and with 2 max-pooling layer. The batch normalization layer is followed by two FC layers and the output layer, respectively. Optimization was performed using the Stochastic Gradient Descent (SGD), with a learning value of 0.001 and a mini-batch size of 32. The maximum number of iterations was set to 1200. To avoid overfitting, the first two FC layers are followed by dropout layers with a 50 percent dropout ratio.
\subsubsection{Convolutional Neural Net 2} 
To develop the Convolutional Neural Net 2 (CNN2) four convolutional layers along with polling for each layer is applied. Each layer is followed by rectified linear units (ReLU) and a max-pooling layer of size 3x3 with stride 2. The first, second, third and fourth layer consists 3x3, 4x16, 3x32, 3x64 filters, respectively. Again each layer is followed by ReLU and a max-pooling layer of size 3x3 with stride 2. Layer 3 is followed by three fully connected layers with 348, 192, and 96 neurons, respectively, which followed by batch normalization layer. For optimization, adaptive moment estimation (ADAM) was utilized, using a learning parameter of 0.0001 and a mini-batch size of 64. The maximum number of iterations was set to 1250.
\subsubsection{Convolutional Neural Net 3}
To develop Convolutional Neural Net 3 (CNN3),five convolutional layers each with a filter size of 3x8, 3x16, 6x16, 3x32, and 12x32 are implemented. Each layer is subjected to polling. The convolutional layer is composed of a number of components, including the filter size, padding, and stride. Each layer is followed by ReLU activation functions and a max-pooling layer with a size of 3x8 and a stride of 2. The batch normalization layer is then followed by four FC layers each having 796,348, 192, and 96 neurons respectively. To prevent overfitting, the first two FC layers are followed by 60 percent dropout layers.
\subsection{Ensemble CNN}
\begin{figure*}[]
    \centering
    \includegraphics{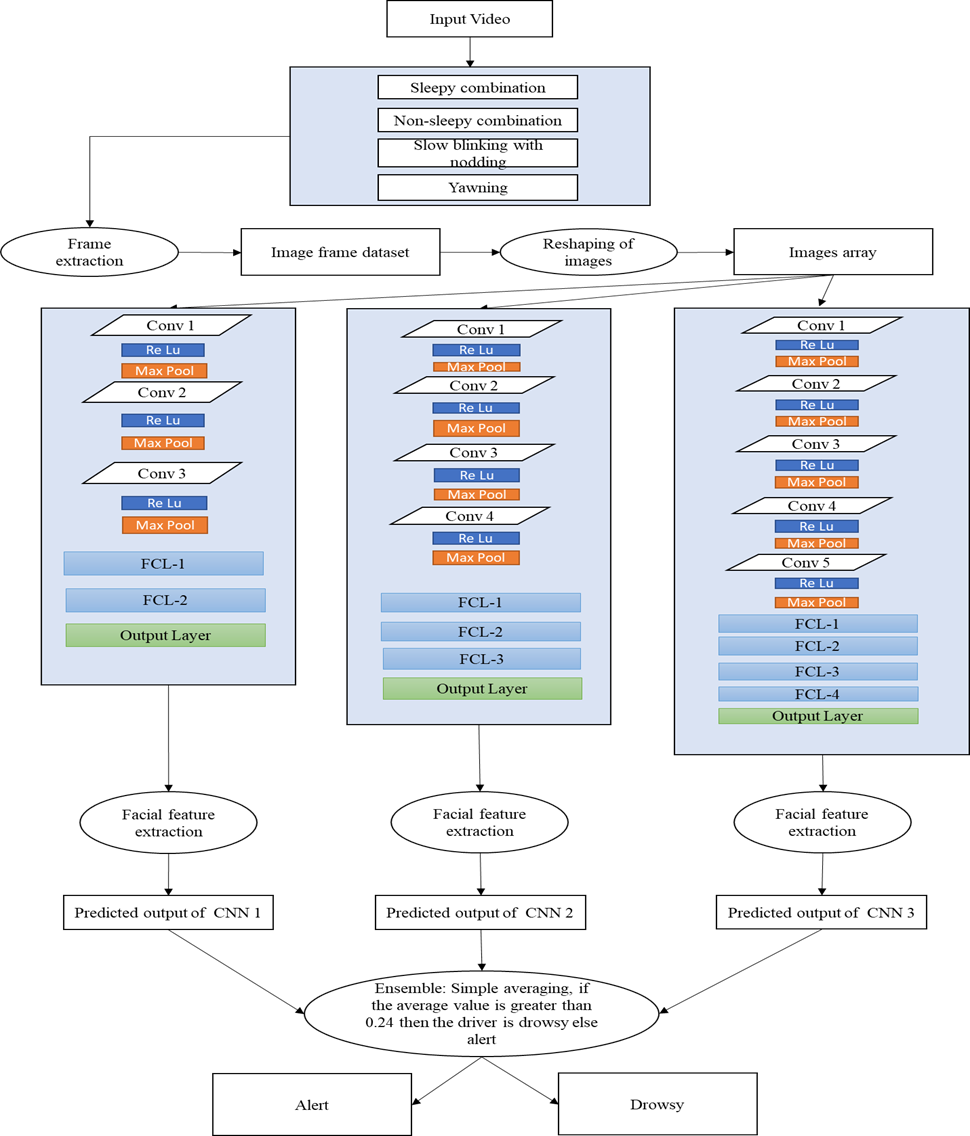}
    \caption{Ensemble CNN structure applied in this study.}
    \label{fig6}
\end{figure*}

Figure~\ref{fig6} shows the Ensemble CNN structure applied in this study. Ensemble-based CNN is used to improve the performance and accuracy of the model. When the individual CNN nets are combined to form the Ensemble net, the model's performance is increased in the form of an F1 score, which can be seen in the Results section. To create a single output, the separate outputs of all models are combined using the simple averaging procedure. Between the non-drowsy and drowsy states, a threshold value is specified. If the result exceeds the threshold value, the driver is considered drowsy; otherwise, non-drowsy. 
Selecting databases is one of the essential steps to build a model. In terms of suitability, usability, and detailing, the data involved should be adequate. Many datasets are available that can be used to frame the models of video-based drowsiness detection. In this study we have used YawDD dataset~\cite{abtahi2014yawdd}, which gathered face videos of a vast number of participants (107 participants). This dataset has been made publicly available to promote further research on drowsiness detection by yawning detection. YawDD consists of 57 male and 50 female participants who were asked to record the video of drivers sitting in the driving seat and pretending to be driving in different light conditions and different facial characteristics like participants wearing glasses and sunglasses, having beards, mustaches, etc. Each participant video lasts for about 15-40 seconds. However, we only evaluated two states/expressions in this experiment: Alert and Drowsy, with labels of 1 and 2 respectively. Some of the most widely used datasets are mentioned in Table~\ref{tab1} along with YawDD dataset in details. 
\begin{table*}[h]
  \caption{Different publicly available driver drowsiness dataset.}
     \centering\resizebox{.8\textwidth}{!}{
    \begin{tabular}{ccccc}\toprule
     No. &	Database&	No. of Subjects&	State&	No. of Images/Videos\\\midrule
     1&	UTA-RLDD~\cite{ghoddoosian2019realistic}&
60&	Alert, & 180 videos\\
&&&low vigilant, drowsy&	\\
2&	NTHU~\cite{weng2016driver}&
36&	Yawning,blinking, &	360 videos\\
&&&nodding, talking,\\ &&&laughing, looking\\
3&	ZJU~\cite{pan2007eyeblink}&
20&	Eye blink&	80 videos\\
4&	ULG~\cite{massoz2016ulg}&
14&	Open/closed eyes&	4157 images\\
5&	YawDD~\cite{abtahi2014yawdd}&
107&normal driving (no talking),&342 videos\\  &&&talking or singing,\\
&&&and yawning while driving\\\bottomrule	 
    \end{tabular}}
  
    \label{tab1}
\end{table*}
At first, we divided the data into three categories: 70\% for training, 15\% for validation, and 15\% for testing. On the YawDD dataset, four CNN-based approaches were utilized to assess overall performance. Table~\ref{dataset} shows the overall data assignment for this study.\\
\begin{table}[h]
\caption{Labelled expressions for YawDD dataset.}
    \centering
    \begin{tabular}{ccc}\toprule
         Expression&	Files&	Labels  \\\midrule
         Alert&	25492&	1\\
         Drowsy&	6733&	2\\
        Total&	32225\\\bottomrule	 
    \end{tabular}
    
    \label{dataset}
\end{table}
\section{Implementation and Results}
\begin{figure*}[]
    \centering
    \includegraphics{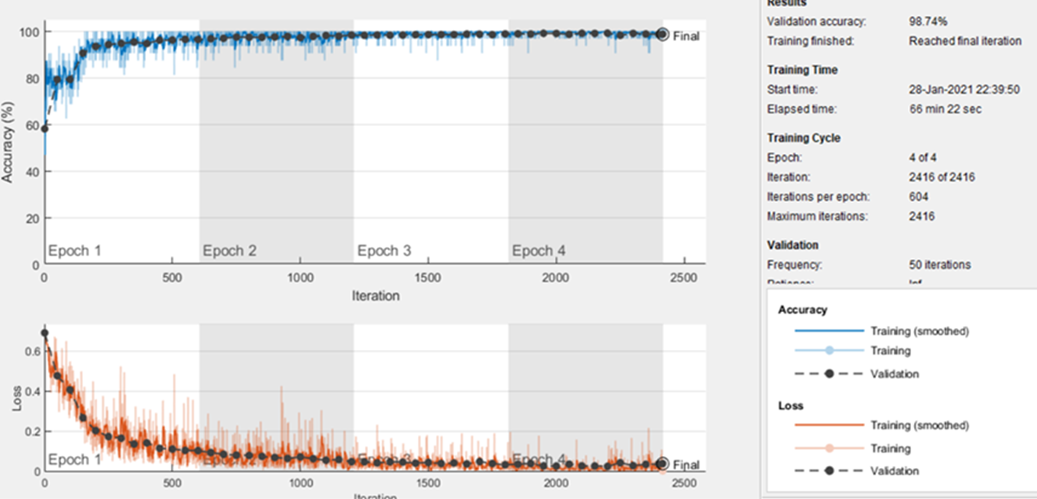}
    \caption{Validation accuracy for convolutional neural net 1.}
    \label{fig8}
\end{figure*}

\begin{figure*}[h!]
    \centering
    \includegraphics[width=\textwidth]{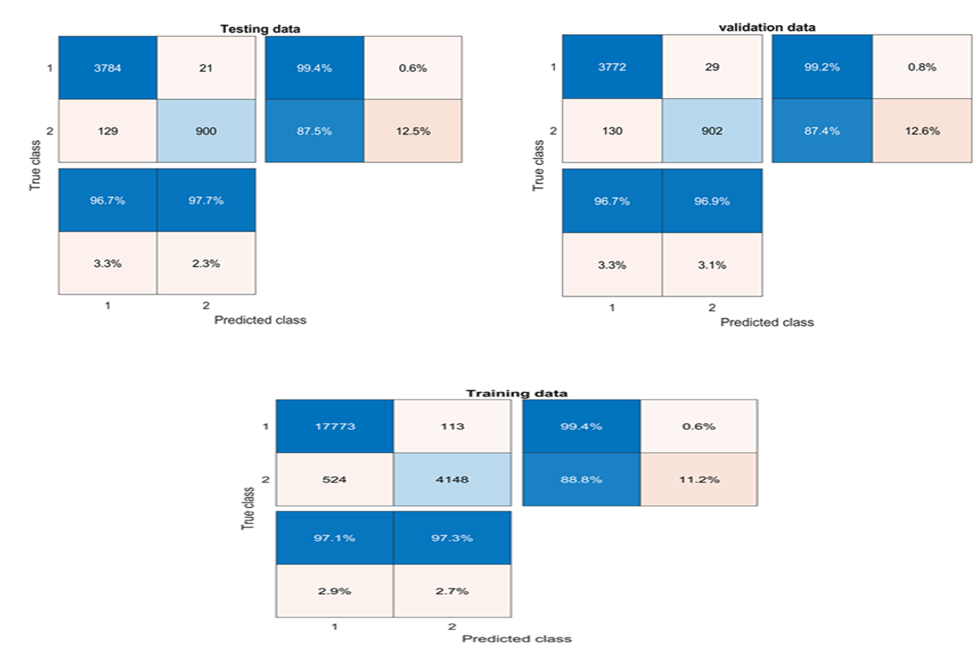}
    \caption{Confusion matrix for testing, training and validation for convolutional neural net 1.}
    \label{fig9}
\end{figure*}

\begin{figure*}[h!]
    \centering
    \includegraphics{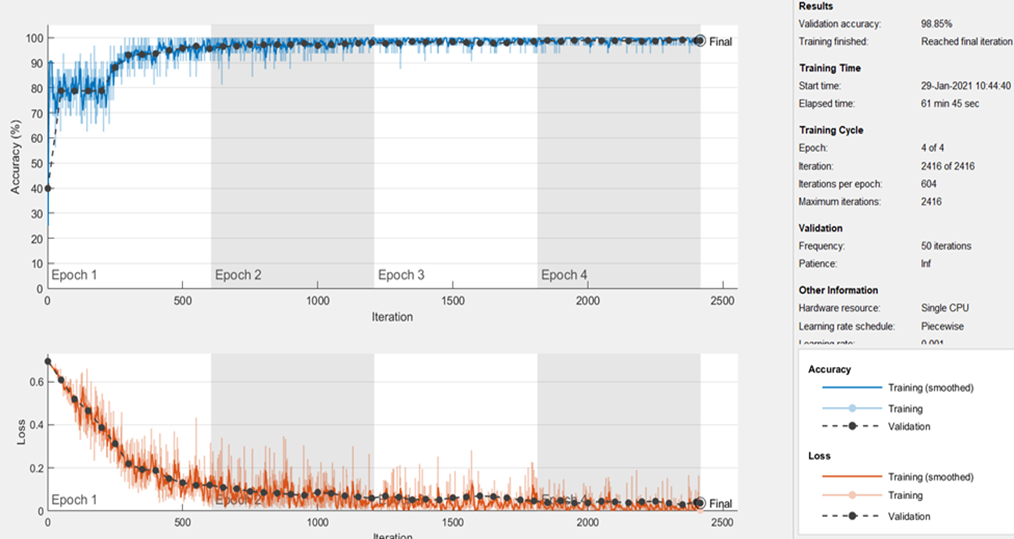}
    \caption{Validation accuracy for convolutional neural net 2.}
    \label{fignet2}
\end{figure*}

\begin{figure*}[h!]
    \centering
    \includegraphics[width=\textwidth]{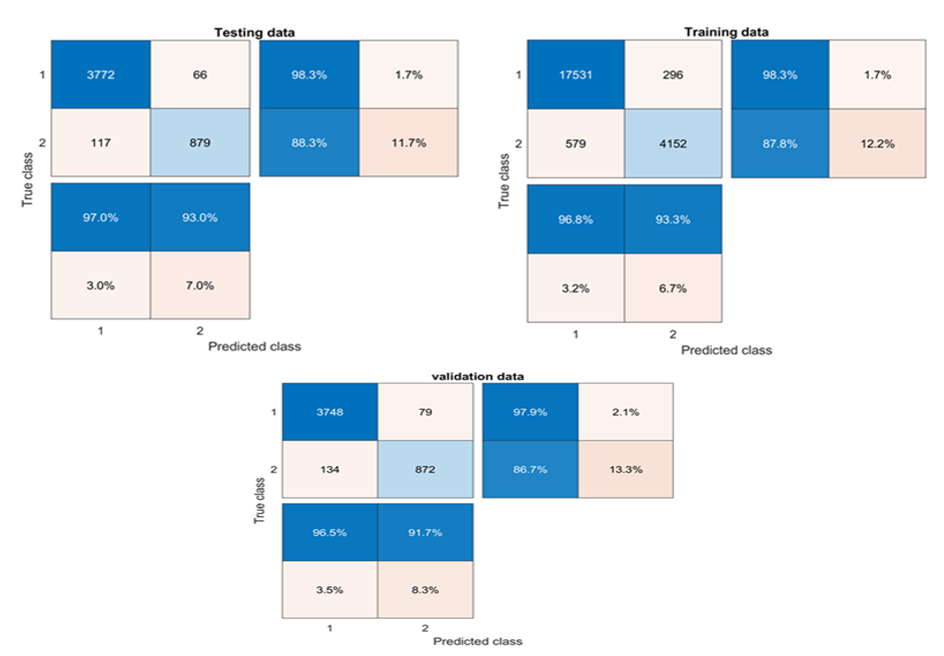}
    \caption{Confusion matrix for testing, training and validation for convolutional neural net 2.}
    \label{figcnn2}
\end{figure*}

Figure~\ref{fig8} shows the validation accuracy for the Convolutional Neural Net 1 (CNN1). The network was trained using stochastic gradient descent with momentum (SGD) with an initial learning rate of 0.001, Minibatch Size of 32, 4 epochs and 2416 iterations. The accuracy for CNN1 is 98.74\%. The elapsed time for CNN1 is 66 min 22 sec.\\
To better visualize the overall performance of prediction, confusion matrices were used. Figure~\ref{fig9} demonstrates that the proposed CNN1 performs nearly identically on training, testing, and validation datasets, effectively reducing the risk of models overfitting.\\
\\
The validation accuracy of the CNN2 is shown in Figure~\ref{fignet2}. Adam, an adaptive learning rate optimization technique with an initial learning rate of 0.001, was used to train the network, with a Minibatch Size of 32,4 epochs and 2416 iterations. CNN2 has a 98.85 percent accuracy rate and took 61 minutes and 45 seconds to complete.\\
Additionally, Figure~\ref{figcnn2} depicts the CNN2 model's confusion matrix for testing, training, and validation data. The model's performance in training and testing is nearly identical, while the CNN2 performance on the validation dataset is slightly worse.

\begin{figure*}
    \centering
    \includegraphics{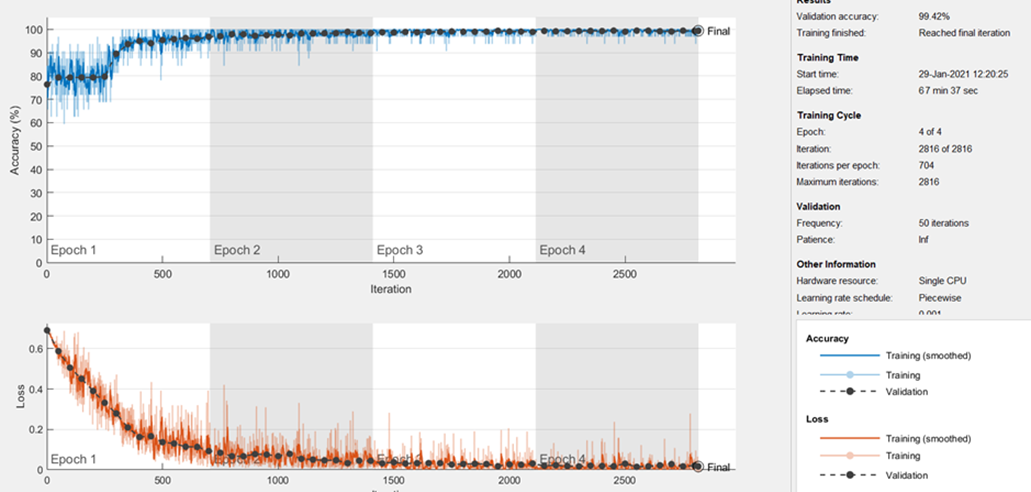}
    \caption{Validation Accuracy for convolutional neural net 3.}
    \label{figcnn3}
\end{figure*}

\begin{figure*}
    \centering
    \includegraphics[width=\textwidth]{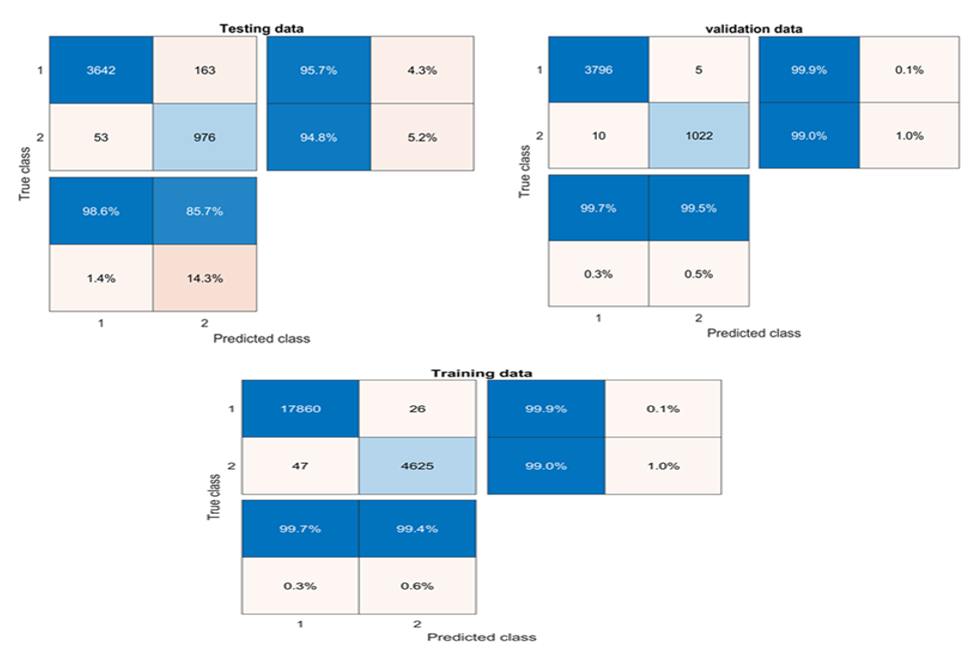}
    \caption{Confusion matrix for testing, training and validation for convolutional neural net 3.}
    \label{figconf3}
\end{figure*}

Here, the validation accuracy of the CNN3 is shown in Figure~\ref{figcnn3}. With an initial learning rate of 0.001, a Minibatch Size of 64,4 epochs, and 2816 iterations, the network was trained using stochastic gradient descent with momentum (SGD). CNN3 has a 99.42 percent accuracy rate and took 67 minutes and 27 seconds to complete.
\\
The confusion matrix for CNN3 models in the training, testing, and validation dataset is shown in Figure~\ref{figconf3}. The best results are seen on the training and validation datasets, whereas the worst results are seen on the testing dataset.
Table~\ref{tabcomparison} displays the Recall, Precision, and F1 Score of CNN1, CNN2, CNN3, and Ensemble Net for the YawDD test dataset, demonstrating that the Ensemble net outperformed individual nets on average in terms of precision, recall and F1 score.\\

\begin{table*}[h!]
\caption{Different CNN models performance on YawDD test dataset.}
    \centering
    \begin{tabular}{ccccccc}\toprule
         	CNN&\multicolumn{2}{c}{Recall}&\multicolumn{2}{c}{Precision}&\multicolumn{2}{c}{F1 Score}\\\midrule
         	&Drowsy	&Alert&	Drowsy&	Alert&	Drowsy&	Alert\\\cmidrule{2-7}
         	Net1&	87.5\%&	99.2\%&	96.9\%&	96.7\%&	0.92&	0.978\\
         	Net2&	88.3\%&	98.2\%&	93\%&	97\%&	0.90&	0.970\\
         	Net3&	85.8\%&	98.2\%&	96.7\%	&97.7\%&	0.912	&0.965\\
         	Ensemble&	90.3\%&	99.5\%&	97\%&	97\%&	0.935&	0.978\\\bottomrule
         & 
    \end{tabular}
    
    \label{tabcomparison}
\end{table*}
\section{Discussion}
Four CNN-based models for detecting driver drowsiness and alertness were proposed and assessed in this work. The results revealed the superiority of Ensemble-based techniques over traditional CNN models employed alone for this purpose. The Ensemble technique outperformed the other CNN models, achieving recall scores of.903 and.995 in the drowsy and alert stages, respectively, compared to the other CNN models. During the alert state, the CNN3 outperformed all other models, achieving a precision score of.977, which was superior to all others. When it came to precision in the drowsy state, Ensemble outperformed all other models with a score of.97, which was higher than any other model. Specifically, when it comes to F1 scores, the Ensemble method and CNN1 both achieved higher scores of.978 in the alert state. The Ensemble model, on the other hand, exceeded all other models in the drowsy state, achieving an F1 score of 0.935 in the meantime.There are some limits to our study, which we believe present immediate potential for future research. These are as follows:\\
\begin{itemize}
    \item The experiment was conducted using only one dataset instead of multiple datasets.
    \item Instead of comparing the performance of the proposed model to that of the referred literature, only four different CNN-based techniques were proposed.
    \item Finally, we did not investigate the compatibility of our proposed models with real-world events in our research. Transnational research should investigate the possibility of bridging that gap as soon as possible in the future, with a higher priority.
\end{itemize}
\section{Conclusion and Future Works}  
In this work, we present four distinct Deep Learning Convolutional Neural Network (CNN)-based models for detecting the driver's drowsiness and alert status using the YawDD dataset. CNN1, CNN2, and CNN3 have an F1-score of 0.972, 0.970, and 0.965, respectively, for alert classification. With an F1 score of 0.978, the proposed Ensemble approaches resulted in improved performance. On the other hand, the F1-scores for CNN1, CNN2, and CNN3 are 0.92, 0.90, and 0.912, respectively, whereas an F1 value of 0.935 indicates improved performance when ensemble methods are used. Thus, based on the findings of our investigation, we can conclude that, of the four offered models, the Ensemble-based CNN technique is the best. Some of our work's shortcomings can be addressed by conducting experiments on significantly imbalanced big data, evaluating our approaches' performance across different datasets, and constructing models with explainable artificial intelligence on a mixed dataset.

\bibliographystyle{unsrt}  
\bibliography{main}

\begin{thebibliography}{10}

\bibitem{ahsan2020face}
Md~Manjurul Ahsan, Yueqing Li, Jing Zhang, Md~Tanvir Ahad, Munshi~Md Yazdan,
  et~al.
\newblock Face recognition in an unconstrained and real-time environment using
  novel bmc-lbph methods incorporates with dji vision sensor.
\newblock {\em Journal of Sensor and Actuator Networks}, 9(4):54, 2020.

\bibitem{gupta2017robust}
Kishor~Datta Gupta, Manjurul Ahsan, Stefan Andrei, and Kazi Md~Rokibul Alam.
\newblock A robust approach of facial orientation recognition from facial
  features.
\newblock {\em BRAIN. Broad Research in Artificial Intelligence and
  Neuroscience}, 8(3):5--12, 2017.

\bibitem{ramzan2019survey}
Muhammad Ramzan, Hikmat~Ullah Khan, Shahid~Mahmood Awan, Amina Ismail, Mahwish
  Ilyas, and Ahsan Mahmood.
\newblock A survey on state-of-the-art drowsiness detection techniques.
\newblock {\em \emph{\textit{IEEE Access}}}, 7:61904--61919, 2019.

\bibitem{abtahi2014yawdd}
Shabnam Abtahi, Mona Omidyeganeh, Shervin Shirmohammadi, and Behnoosh Hariri.
\newblock Yawdd: A yawning detection dataset.
\newblock In {\em \emph{\textit{Proceedings of the 5th ACM multimedia systems
  conference}}}, pages 24--28, 2014.

\bibitem{weng2016driver}
Ching-Hua Weng, Ying-Hsiu Lai, and Shang-Hong Lai.
\newblock Driver drowsiness detection via a hierarchical temporal deep belief
  network.
\newblock In {\em \emph{\textit{Asian Conference on Computer Vision}}}, pages
  117--133. Springer, 2016.

\bibitem{revelo2019human}
Adriana Revelo, Robin {\'A}lvarez, and Felipe Grijalva.
\newblock Human drowsiness detection in real time, using computer vision.
\newblock In {\em \emph{\textit{2019 IEEE Fourth Ecuador Technical Chapters
  Meeting (ETCM)}}}, pages 1--6. IEEE, 2019.

\bibitem{pan2007eyeblink}
Gang Pan, Lin Sun, Zhaohui Wu, and Shihong Lao.
\newblock Eyeblink-based anti-spoofing in face recognition from a generic
  webcamera.
\newblock In {\em \emph{\textit{2007 IEEE 11th international conference on
  computer vision}}}, pages 1--8. IEEE, 2007.

\bibitem{ahsan2021detection}
Md~Manjurul Ahsan, Redwan Nazim, Zahed Siddique, and Pedro Huebner.
\newblock Detection of covid-19 patients from ct scan and chest x-ray data
  using modified mobilenetv2 and lime.
\newblock In {\em Healthcare}, volume~9, page 1099. Multidisciplinary Digital
  Publishing Institute, 2021.

\bibitem{ahsan2021detecting}
Md~Manjurul Ahsan, Md~Tanvir Ahad, Farzana~Akter Soma, Shuva Paul, Ananna
  Chowdhury, Shahana~Akter Luna, Munshi Md~Shafwat Yazdan, Akhlaqur Rahman,
  Zahed Siddique, and Pedro Huebner.
\newblock Detecting sars-cov-2 from chest x-ray using artificial intelligence.
\newblock {\em Ieee Access}, 9:35501--35513, 2021.

\bibitem{ahsan2020deep}
Md~Manjurul Ahsan, Tasfiq E~Alam, Theodore Trafalis, and Pedro Huebner.
\newblock Deep mlp-cnn model using mixed-data to distinguish between covid-19
  and non-covid-19 patients.
\newblock {\em Symmetry}, 12(9):1526, 2020.

\bibitem{ahsan2020covid}
Md~Manjurul Ahsan, Kishor~Datta Gupta, Mohammad~Maminur Islam, Sajib Sen,
  Md~Rahman, Mohammad Shakhawat~Hossain, et~al.
\newblock Covid-19 symptoms detection based on nasnetmobile with explainable ai
  using various imaging modalities.
\newblock {\em Machine Learning and Knowledge Extraction}, 2(4):490--504, 2020.

\bibitem{manjurul2021machine}
Md~Manjurul~Ahsan and Zahed Siddique.
\newblock Machine learning-based heart disease diagnosis: A systematic
  literature review.
\newblock {\em arXiv e-prints}, pages arXiv--2112, 2021.

\bibitem{zhou2012image}
Dengwen Zhou, Xiaoliu Shen, and Weiming Dong.
\newblock Image zooming using directional cubic convolution interpolation.
\newblock {\em \emph{\textit{IET Image Processing}}}, 6(6):627--634, 2012.

\bibitem{zhang2015multimodal}
Wei Zhang, Youmei Zhang, Lin Ma, Jingwei Guan, and Shijie Gong.
\newblock Multimodal learning for facial expression recognition.
\newblock {\em \emph{\textit{Pattern Recognition}}}, 48(10):3191--3202, 2015.

\bibitem{xie2018real}
Yongquan Xie, Kexun Chen, and Yi~Lu Murphey.
\newblock Real-time and robust driver yawning detection with deep neural
  networks.
\newblock In {\em \emph{\textit{2018 IEEE Symposium Series on Computational
  Intelligence (SSCI)}}}, pages 532--538. IEEE, 2018.

\bibitem{krizhevsky2017imagenet}
Alex Krizhevsky, Ilya Sutskever, and Geoffrey~E Hinton.
\newblock Imagenet classification with deep convolutional neural networks.
\newblock {\em \emph{\textit{Communications of the ACM}}}, 60(6):84--90, 2017.

\bibitem{parkhi2015deep}
Omkar~M Parkhi, Andrea Vedaldi, and Andrew Zisserman.
\newblock Deep face recognition.
\newblock 2015.

\bibitem{donahue2015long}
Jeffrey Donahue, Lisa Anne~Hendricks, Sergio Guadarrama, Marcus Rohrbach,
  Subhashini Venugopalan, Kate Saenko, and Trevor Darrell.
\newblock Long-term recurrent convolutional networks for visual recognition and
  description.
\newblock In {\em \emph{\textit{Proceedings of the IEEE conference on computer
  vision and pattern recognition}}}, pages 2625--2634, 2015.

\bibitem{he2016deep}
Kaiming He, Xiangyu Zhang, Shaoqing Ren, and Jian Sun.
\newblock Deep residual learning for image recognition.
\newblock In {\em \emph{\textit{Proceedings of the IEEE conference on computer
  vision and pattern recognition}}}, pages 770--778, 2016.

\bibitem{park2016driver}
Sanghyuk Park, Fei Pan, Sunghun Kang, and Chang~D Yoo.
\newblock Driver drowsiness detection system based on feature representation
  learning using various deep networks.
\newblock In {\em \emph{\textit{Asian Conference on Computer Vision}}}, pages
  154--164. Springer, 2016.

\bibitem{ba2013deep}
Lei~Jimmy Ba and Rich Caruana.
\newblock Do deep nets really need to be deep?
\newblock {\em \emph{\textit{arXiv preprint arXiv:1312.6184}}}, 2013.

\bibitem{mohanraj2019ensemble}
V~Mohanraj, S~Sibi Chakkaravarthy, and V~Vaidehi.
\newblock Ensemble of convolutional neural networks for face recognition.
\newblock In {\em Recent Developments in Machine Learning and Data Analytics},
  pages 467--477. Springer, 2019.

\bibitem{choi2018driver}
Hyung-Tak Choi, Moon-Ki Back, and Kyu-Chul Lee.
\newblock Driver drowsiness detection based on multimodal using fusion of
  visual-feature and bio-signal.
\newblock In {\em \emph{\textit{2018 International Conference on Information
  and Communication Technology Convergence (ICTC)}}}, pages 1249--1251. IEEE,
  2018.

\bibitem{ukwuoma2019deep}
Chiagoziem~C Ukwuoma and Chen Bo.
\newblock Deep learning review on drivers drowsiness detection.
\newblock In {\em \emph{\textit{2019 4th Technology Innovation Management and
  Engineering Science International Conference (TIMES-iCON)}}}, pages 1--5.
  IEEE, 2019.

\bibitem{nikolskaia2019prototype}
Kseniia Nikolskaia, Vladislav Bessonov, Artem Starkov, and Aleksey Minbaleev.
\newblock Prototype of driver fatigue detection system using convolutional
  neural network.
\newblock In {\em \emph{\textit{2019 International Conference" Quality
  Management, Transport and Information Security, Information
  Technologies"(IT\&QM\&IS)}}}, pages 82--86. IEEE, 2019.

\bibitem{delbiaggio2017comparison}
Nicolas Delbiaggio.
\newblock A comparison of facial recognition’s algorithms.
\newblock 2017.

\bibitem{srivastava2017survey}
Ankit Srivastava, Suraj Mane, Aaditya Shah, Nirmit Shrivastava, and Bhushan
  Thakare.
\newblock A survey of face detection algorithms.
\newblock In {\em \emph{\textit{2017 International Conference on Inventive
  Systems and Control (ICISC)}}}, pages 1--4. IEEE, 2017.

\bibitem{yuen2017occluded}
Kevan Yuen and Mohan~M Trivedi.
\newblock An occluded stacked hourglass approach to facial landmark
  localization and occlusion estimation.
\newblock {\em \emph{\textit{IEEE Transactions on Intelligent Vehicles}}},
  2(4):321--331, 2017.

\bibitem{mandal2016towards}
Bappaditya Mandal, Liyuan Li, Gang~Sam Wang, and Jie Lin.
\newblock Towards detection of bus driver fatigue based on robust visual
  analysis of eye state.
\newblock {\em \emph{\textit{IEEE Transactions on Intelligent Transportation
  Systems}}}, 18(3):545--557, 2016.

\bibitem{savacs2018real}
Burcu~K{\i}r Sava{\c{s}} and Ya{\c{s}}ar Becerikli.
\newblock Real time driver fatigue detection based on svm algorithm.
\newblock In {\em \emph{\textit{2018 6th International Conference on Control
  Engineering \& Information Technology (CEIT)}}}, pages 1--4. IEEE, 2018.

\bibitem{kumar2018driver}
Ashish Kumar and Rusha Patra.
\newblock Driver drowsiness monitoring system using visual behaviour and
  machine learning.
\newblock In {\em \emph{\textit{2018 IEEE Symposium on Computer Applications \&
  Industrial Electronics (ISCAIE)}}}, pages 339--344. IEEE, 2018.

\bibitem{lyu2018long}
Jie Lyu, Zejian Yuan, and Dapeng Chen.
\newblock Long-term multi-granularity deep framework for driver drowsiness
  detection.
\newblock {\em \emph{\textit{arXiv preprint arXiv:1801.02325}}}, 2018.

\bibitem{boubenna2018image}
Hadjer Boubenna and Dohoon Lee.
\newblock Image-based emotion recognition using evolutionary algorithms.
\newblock {\em \emph{\textit{Biologically inspired cognitive architectures}}},
  24:70--76, 2018.

\bibitem{ghoddoosian2019realistic}
Reza Ghoddoosian, Marnim Galib, and Vassilis Athitsos.
\newblock A realistic dataset and baseline temporal model for early drowsiness
  detection.
\newblock In {\em \emph{\textit{Proceedings of the IEEE/CVF Conference on
  Computer Vision and Pattern Recognition Workshops}}}, pages 0--0, 2019.

\bibitem{massoz2016ulg}
Quentin Massoz, Thomas Langohr, Cl{\'e}mentine Fran{\c{c}}ois, and Jacques~G
  Verly.
\newblock The ulg multimodality drowsiness database (called drozy) and examples
  of use.
\newblock In {\em \emph{\textit{2016 IEEE Winter Conference on Applications of
  Computer Vision (WACV)}}}, pages 1--7. IEEE, 2016.

\end{thebibliography}

\end{document}